\documentclass{article}

\usepackage{ijcai13}

\newtheorem {definition}{Definition}
\newtheorem {proposition}{Proposition}

\newtheorem {theorem}{Theorem}

\newtheorem {example}{Example}

\usepackage{amsmath}
\usepackage{amssymb}
\usepackage{rotating}

\begin{document}

\title{Nested Aggregates in Answer Sets \\ {\small Application to a Priori Optimization } }

\author{ Emad Saad \\
emsaad@gmail.com
}

\maketitle

\begin{abstract}

We allow representing and reasoning in the presence of nested multiple aggregates over multiple variables and nested multiple aggregates over functions involving multiple variables in answer sets, precisely, in answer set optimization programming and in answer set programming. We show the applicability of the answer set optimization programming with nested multiple aggregates and the answer set programming with nested multiple aggregates to the Probabilistic Traveling Salesman Problem, a fundamental {\em a priori optimization} problem in Operation Research.
\end{abstract}

\section{Introduction}

A fruitful approach to aggregates in answer set programming has been presented in \cite{Recur-aggr} that allows to represent and reason in the presence of aggregate function which are defined through aggregates atoms that are allowed to be recursive. The aggregate functions defined in \cite{Recur-aggr}, and hence the aggregate atoms defined over them, are allowed to be monotone, anti-monotone and non-monotone which is clearly appropriate for non-monotonic reasoning. The answer set semantics with aggregates defined in \cite{Recur-aggr} have shown to be appropriate for many interesting problems in answer set programming that require aggregates over some numerical criteria imposed by the problems and arise in many domains.

On the other hand aggregates have been considered in answer set optimization programming described in \cite{Saad_ASOG} that extended the answer set optimization programs \cite{ASO} to allow representing and {\em declaratively} solving multi-objective optimization problems in an answer set programming framework. The answer set optimization with aggregates framework of \cite{Saad_ASOG} have shown to be able to find Nash equilibrium in strategic games with any number of players and with any number of strategies, which is a multi-objective optimization problem with multiple conflicting goals.

However the aggregates defined in the answer set programming framework in \cite{Recur-aggr} allow a {\em single} aggregation over a {\em single} variable which limits its applicability to many interesting problems, especially the problems that require aggregations over a function with multiple variables. And hence, more generally limits its applicability to the problems that require nested multiple aggregations over multiple variables and the problems that require nested multiple aggregations over functions with multiple variables.

On the other hand, although the aggregates defined in answer set optimization programming \cite{Saad_ASOG}, is similar to the aggregates defined in the answer set programming \cite{Recur-aggr}, allows a {\em single} aggregation over a {\em single} variable, the answer set optimization programming in \cite{Saad_ASOG} is still capable of solving some interesting multi-objective optimization problems like the one arises from finding Nash equilibrium in strategic games. However, the answer set optimization programming of \cite{Saad_ASOG} is still incapable of representing and reasoning about many interesting optimization problems that require optimization over objective functions that involve nested multiple aggregation over functions that contain multiple variables. This implies minimization or maximization over nested multiple aggregations over functions with multiple variables or nested multiple aggregations over multiple variables.

In this paper we generalize both answer set optimization programming \cite{Saad_ASOG} and answer set programming \cite{Recur-aggr} with nested multiple aggregates to allow representing and reasoning in the presence of nested multiple aggregation over multiple variables and nested multiple aggregation over functions with multiple variables. We show the applicability of the nested multiple aggregates in answer sets to the {\em Probabilistic Traveling Salesman Problem} (PTSP), a fundamental stochastic optimization problem in Operation Research \cite{PTSP}, whose objective function involving nested multiple aggregations over a function with multiple variables.

Probabilistic Traveling Salesman Problem is a {\em priori optimization} problem for which we want to find {\em a priori a tour} with minimum (expected) length through a set of $n$ points, where only $k$ points out of $n$ points ($0 \leq k \leq n$) at any given instance of the problem must be visited, where the number $k$ is chosen at random with a known probability distribution, that is based on the probability distribution of visiting each point in the set of $n$ points. In addition, the $k$ points chosen to being visited, in a given instance of the problem, has to be visited in the same order as they appear in the a priori tour.

Formally, the Probabilistic Traveling Salesman Problem is defined as follows. A Probabilistic Traveling Salesman Problem (PTSP), $G$, is a tuple of the form $G = \langle V, E, D, P \rangle$, where $\langle V, E \rangle$ is a graph with a set of $n$ nodes, $V$, and a set of arcs, $E$, connecting the nodes in $V$, $D$ is an $n \times n$ (complete) matrix (the distance matrix) that specifies the distance (cost), $d_{ij}$, incurred by traveling from a node (location) $i$ to a node (location) $j$, $P = \{p_1, p_2, \ldots, p_n\}$ is a probability distribution over the nodes in $V$, where $p_i$ is the probability of a node $i$ must be visited. In other words, the nodes in $V$ represent the set of locations that need to be visited, arcs in $E$ represent the roads connecting the locations, the values $d_{ij}$ in $D$ represent the distance traveling from location $i$ to location $j$, and the probabilities $p_i$ in $P$ represent the probability that a location $i$ must be visited. The objective is to find {\em a priori tour} $t$, a Hamiltonian circuit of $G$, with the minimum expected length (cost). By indexing the nodes of a tour $t$ by their order of appearance, we present the tour, $t$, as the sequence of nodes $t = (0, 1, 2, 3, 4,  \ldots, n, 0 )$, where $0$ is the depot or the starting point from which the tour always starts and ends. It has the index $0$ and with probability $p_0 = 1$, since it must be always visited. Therefore, the expected length, $L_t$ of a tour $t$, denoted by $E[L_t]$, is given by:
\begin{equation}
E[L_t] = \sum_{i = 0}^{n} \sum_{j = i + 1}^{n + 1}
 d_{ij} . p_i . p_j .
\left( \prod_{k = i+1}^{j - 1} (1 - p_k) \right)
\label{eq:expected}
\end{equation}
where $i, j, k$ are indices of nodes within a tout $t$ (not the nodes themselves) and $n + 1 = 0$. Given that $\cal T$ is the set of all tours in $G$, the objective is to find a tour $t$ in $\cal T$ with the minimum expected length $E[L_t]$, i.e., solve the a priori optimization problem
\[
\min_{t \in {\cal T}} E[L_t].
\]

In this paper we develop syntax and semantics for both answer set optimization programs and answer set programs to allow representing and reasoning in the presence of nested multiple aggregates over multiple variables and nested multiple aggregates over functions involving multiple variables. We show the applicability of both the answer set optimization programs with nested multiple aggregates and the answer set programs with nested multiple aggregates to the fundamental stochastic optimization Probabilistic Traveling Salesman Problem, {\em a priori optimization} problem in Operation Research \cite{PTSP}. In addition, we show that with minor modification to the probabilistic traveling salesman problems representation in answer set optimization programs with nested multiple aggregates and in answer set programs with nested multiple aggregates, a corresponding classical traveling salesman problems can be intuitively represented and solved in both frameworks. We prove that the presented answer set optimization programs with nested multiple aggregates modify and generalize the answer set optimization programs with a single aggregate over a single variable proposed in \cite{Saad_ASOG} as well as a generalization of the answer set optimization programs described in \cite{ASO}. In addition, we prove that the presented answer set programs with nested multiple aggregates generalize the answer set programs with a single aggregate over a single variable described in \cite{Recur-aggr} as well as a generalization of the original answer set programs presented in \cite{Gelfond_B}.

\section{Nested Aggregates Disjunctive Logic Programs}

In this section we present the syntax and the answer set semantics of disjunctive logic programs (a form of answer set programming) with nested multiple aggregates, denoted by NDLP. The syntax and the answer set semantics of NDLP generalize and modifies the syntax and semantics of disjunctive logic programs with a single aggregate over a single variable, DLP$^{\cal A}$, presented in \cite{Recur-aggr}.

\subsection{NDLP Programs Syntax}

Let $\cal L$ be a first-order language with finitely many predicate symbols, function symbols, constants, and
infinitely many variables. A term is a constant, a variable or a function. An atom is a predicate in $B_{\cal L}$, where $B_{\cal L}$ is the Herbrand base of ${\cal L}$. The Herbrand universe of ${\cal L}$ is denoted by $U_{\cal L}$. Non-monotonic negation or the negation as failure is denoted by $not$.

An expression of the form $\{ F \; | \; C \}$ is called a symbolic set, where $F$ is a single variable or a function $F = g(X_1, \ldots, X_n)$, where $X_1, \ldots, X_n$ are variables or functions and $C$ is a conjunction of atoms, negation of atoms, aggregate atoms, and negation of aggregate atoms (defined below). However, a ground set is a set of pairs of the form $\langle F_G \;|\; C_G \rangle$ such that $F_G$ is a constant and $C_G$ is a conjunction of ground atoms, negated ground atoms, ground aggregate atoms, and negated ground aggregate atoms (defined below). A ground set or a symbolic set is called a set term. We say $f(S)$ is an aggregate function if $f$ is an aggregate function symbol and $S$ is a set term, where $f \in \{ min, max, count, sum, times \}$. If $f(S)$ is an aggregate function and $T$ is a constant, a variable or a function term called guard, then we say $f(S) \prec T$ is an aggregate atom, where $\prec \in \{=, \neq, <, >, \leq, \geq \}$.

Observe that, from the definition of aggregate functions and aggregate atoms, it can be seen that an aggregate function $f(S)$ in an aggregate atom $f(S) \prec T$ can be defined in terms of other aggregate atoms $f'(S') \prec T'$ and hence in terms of other aggregate functions $f'(S')$. This means multiple level of nested aggregate functions calculation is achieved by allowing aggregate functions of the aggregate atoms to be defined in terms of other aggregate atoms. This is accomplished by allowing the conjunction in the definition of the symbolic set $S$ of the aggregate function $f(S)$ in the aggregate atom $f(S) \prec T$ to contain other aggregate atoms as constituents. This can be illustrated by the following example.

\begin{example} Assume that we want to represent the double summation
\begin{equation}
\sum_{i = 1}^{n} \sum_{j = 1}^{m} (i+j)
\label{eq:summation}
\end{equation}
as an aggregate function, $f(S)$, and assign the result to a variable called $X$ to form the aggregate atom $f(S) = X$. Assume also that the possible values of $i,j$ are represented by the predicate $d(I,J)$. Therefore, the double summation (\ref{eq:summation}) can be represented as the aggregate atom $f(S) = X$ that is defined as:
\begin{eqnarray}
sum \; \{ \; A \; | \; 1 \leq I \leq n, \notag \\ sum \; \{ \; I+J \; | \; d(I,J), \;  1 \leq J \leq m, \;   \; \} \; = \; A \; \} \; = \;  X
\label{atom:summation}
\end{eqnarray}
where $ S = \{ \; A \; | \; 1 \leq I \leq n, \; sum \; \{ \; I+J \; | \; d(I,J), \;  1 \leq J \leq m, \;   \; \} \; = \; A \; \}$.
\label{ex:summation}
\end{example}
Notice that there are two instances of the aggregate function $sum$ in the aggregate atom representation of the double summation (\ref{eq:summation}). The {\em outer} instance of the aggregate function $sum$ and the {\em inner} instance of the aggregate function $sum$. The inner instance of the aggregate function $sum$ is defined inside the outer instance of the aggregate function $sum$.

Moreover, notice that the variable $J$ occurs only in the inner instance of the aggregate function $sum$. Whereas the variable $I$ occurs in both the outer and the inner instances of the aggregate function $sum$. This means that the variable $J$ is seen only by the inner aggregate function $sum$, but not seen by the outer aggregate function $sum$, and $I$ is seen by both the outer and the inner aggregate function $sum$. This implies that the occurrence of $I$ in the outer $sum$ hides the occurrence of $I$ in the inner $sum$, which means that any substitution of $I$ to a constant in both the inner and the outer $sum$ is made by the outer $sum$. This also implies that the {\em scope} of the variable $J$ is the inner $sum$ while the {\em scope} of the variable $I$ is both the outer and the inner $sum$, i.e., the entire symbolic set of the outer sum. This motivates the following definition of local variables to aggregate functions.

\begin{definition}
Let $f(S)$ be an aggregate function. A variable, $X$, is a local variable to $f(S)$ if and only if $X$ appears in $S$ and $X$ does not appear in any aggregate function that is outer to $f(S)$ or in the NDLP rule that contains $f(S)$.
\label{def:local}
\end{definition}
Definition (\ref{def:local}) specifies that every aggregate function $f(S)$ has its own set of local variables. For example, for the aggregate atom representation of the double summation described in Example (\ref{ex:summation}), the variable $I$ is local variable to the outer aggregate function $sum$ while the variable $J$ is local variable to the inner aggregate function $sum$.

\begin{definition}
A global variable is a variable that is not a local variable.
\end{definition}

\begin{definition} An NDLP program, $\Pi$, is a set of NDLP rules of the form
\begin{equation}
a_1 \vee a_2 \vee \ldots \vee a_k \leftarrow a_{k+1}, \ldots, a_m, not\; a_{m+1},\ldots, not\;a_{n} \label{rule:DLPN}
\end{equation}
where $a_1, a_2, \ldots, a_k$ are atoms and $a_{k+1}, \ldots, a_{n}$ are atoms or aggregate atoms.
\end{definition}
Let $r$ be an  NDLP rule of the form (\ref{rule:DLPN}). We use $head(r) = a_1 \vee a_2 \vee \ldots \vee a_k$ and $body(r) = a_{k+1},\ldots, a_m, not\; a_{m+1},\ldots, not\;a_{n}$.

\subsection{NDLP Programs Semantics}

\begin{definition}
The {\em local ground instantiation} of a symbolic set $S = \{ F  \; | \; C \}$ is the set of all local ground pairs of the form $\langle \theta\; (F) \; | \;  \theta \; (C) \rangle$, where $\theta$ is a substitution of every local variable appearing in $S$ to a constant from $U_{\cal L}$.
\end{definition}

\begin{definition}
Let $S$ be a symbolic set. Then, the ground instantiation of $S$ is the local ground instantiation of $S$, then followed by the local ground instantiation of every symbolic set, $S'$, appearing in $S$, then followed by the local ground instantiation of every symbolic set, $S''$, appearing in $S'$, then etc.
\end{definition}

\begin{example} Consider the grounding of the aggregate function representation of the double summation presented in Example (\ref{ex:summation}) and described by (\ref{atom:summation}), where $n = m = 2$ and with the following facts added.
\[
\begin{array}{c}
d(1,1). \hspace{0.5cm} d(1,2). \hspace{0.5cm} d(2,1). \hspace{0.5cm} d(2,2). \hspace{0.5cm} e(3,1).
\end{array}
\]
Since the aggregate function in (\ref{atom:summation}) contains two levels of nesting of the aggregate function $sum$, then the grounding is achieved in two steps. The first step is described as follows, where $A_1$, $A_2$, and $A_3$ are variables act as place holders that are replaceable by constants.
\[
\begin{array}{c}
sum \{
\qquad \; \langle A_1  \; | \; (1 \leq 1 \leq 2), \\ sum \{ \; 1+J \; | \; d(1,J), (1 \leq J \leq 2)  \}  =  A_1  \rangle,
\langle A_2  \; | \; (1 \leq 2 \leq 2), \\ sum \{ \; 2+J \; | \; d(2,J), (1 \leq J \leq 2)  \}  =  A_2  \rangle ,
\langle A_3  \; | \; (1 \leq 3 \leq 2), \\ sum \{ \; 3+J \; | \; d(3,J), (1 \leq J \leq 2)  \}  =  A_3  \rangle
\qquad
\}
\end{array}
\]
followed by the second step which is described as:
\[
\begin{array}{c}
sum \; \{ \qquad
\langle \; A_1  \; | \; (1 \leq 1 \leq 2), \\
sum \; \{ \;
\langle 1+1 | d(1,1), (1 \leq 1 \leq 2) \rangle, \langle 1+2 | d(1,2), (1 \leq 2 \leq 2) \rangle, \\ \langle 1+3 | d(1,3), (1 \leq 3 \leq 2) \rangle
\; \}  =  A_1
\; \rangle , \\
\\
\qquad
\langle \; A_2  \; | \; (1 \leq 2 \leq 2), \\
sum \; \{ \;
\langle 2+1 | d(2,1), (1 \leq 1 \leq 2) \rangle, \langle 2+2 | d(2,2), (1 \leq 2 \leq 2) \rangle, \\ \langle 2+3 | d(2,3), (1 \leq 3 \leq 2) \rangle
\; \}  =  A_2
\; \rangle , \\
\\
\qquad
\langle \; A_3  \; | \; (1 \leq 3 \leq 2), \\
sum \; \{ \;
\langle 3+1 | d(3,1), (1 \leq 1 \leq 2) \rangle, \langle 3+2 | d(3,2), (1 \leq 2 \leq 2) \rangle, \\ \langle 3+3 | d(3,3), (1 \leq 3 \leq 2) \rangle
\; \}  =  A_3
\; \rangle  \\
\\
\qquad
\qquad
\}
\end{array}
\]
\label{ex:ground-summation}
\end{example}

\begin{definition} A ground instantiation of an NDLP rule, $r$, is the replacement of each global variable appearing in $r$ to a constant from $U_{\cal L}$, then followed by the ground instantiation of every symbolic set, $S$, appearing in $r$.
\end{definition}
%
%
The ground instantiation of an NDLP program, $\Pi$, is the set of all possible ground instantiations of every NDLP rule, $r$, in  $\Pi$.

Let $\mathbb{X}$ be a set of objects. Then, we use $\overline{2}^\mathbb{X}$ to denote the set of all multisets over elements in $\mathbb{X}$. The semantics of the aggregate functions $min$, $max$, $count$, $sum$, and $times$ are defined by the mappings; $min, max: (\overline{2}^{\mathbb{R}} - \emptyset) \rightarrow \mathbb{R}$; $count : \overline{2}^{U_{\cal L}}  \rightarrow \mathbb{N}$, $sum : \overline{2}^{\mathbb{R}} \rightarrow \mathbb{R}$, and $times: \overline{2}^{\mathbb{R}} \rightarrow \mathbb{R}$, where $\mathbb{R}$ is the set
of all real numbers, $\mathbb{N}$ is the set of all natural numbers, and $U_{\cal L}$ is the Herbrand universe. The application of $sum$ and $times$ on the empty multiset returns zero and one respectively. The application of $count$ on the empty multiset returns zero. However, the application of $max$ and, $min$ on the empty multiset is undefined. Let $\bot$ be a symbol that does not occur in any NDLP program.

\begin{definition} An interpretation is a subset of the Herbrand base ${\cal B_L}$.
\end{definition}
An atom, $a$, is true (satisfied) with respect to an interpretation, $I$, if $a$ belongs to $I$; but it is false (unsatisfied) otherwise. The negation of an atom, $not \; a$, is true (satisfied) with respect to $I$ if $a$ does not belong to $I$; but it is false (unsatisfied) otherwise. Similarly, the evaluation of an aggregate function, and hence the truth valuation of an aggregate atom, are established with respect to a given interpretation, $I$, as described by the following definitions.

\begin{definition} Let $f(S)$ be a ground aggregate function and $I$ be an interpretation. Then, we define $S_I$ to be the multiset constructed from elements in $S$, where $S_I = \{\!\!\{F_G \; | \; \langle F_G \: | \: C_G \rangle \in S \wedge$ $C_G$ is true in $I \}\!\!\}$.
\end{definition}

\begin{definition} Let $f(S)$ be a ground aggregate function and $I$ be an interpretation. Then, the evaluation of $f(S)$ with respect to $I$ is, $f(S_I)$, the result of the application of $f$ to $S_I$. $f(S_I) = \bot$ if $S_I$ is not in the domain of $f$.
\end{definition}

\begin{example} Let $I = \{ d(1,1), d(1,2), d(2,1), d(2,2), e(3,1) \}$ be an interpretation. Thus, the evaluation of the ground aggregate function representing the double summation in Example (\ref{ex:ground-summation}) is evaluated w.r.t. $I$ as follows in two steps. Considering only the relevant possible values of the variables $A_1$, $A_2$, and $A_3$ in Example (\ref{ex:ground-summation}), the first step is to evaluate the inner instance of the aggregate function $sum$ as follows:
\[
\begin{array}{c}
sum \; \{
\langle \; 5  \; | \; (1 \leq 1 \leq 2), sum \; \{ \;
2, 3 \; \}  =  5
\; \rangle ,
\langle \; 7  \; | \; (1 \leq 2 \leq 2), \\sum \; \{ \; 3, 4 \; \}  =  7 \; \rangle
 \}
\end{array}
\]
followed by evaluating the second instance of the aggregate function $sum$ as:
\[
\begin{array}{c}
sum \; \{ \; 5 , 7 \;  \} = 12 = \sum_{i = 1}^ 2 \sum_{j = 1}^2 (i+j)
\end{array}
\]
\label{ex:evaluation-summation}
\end{example}
\begin{definition} Let $f(S) \prec T$ be a ground aggregate atom and $I$ be an interpretation. Then, $f(S) \prec T$ is true (satisfied) with respect to $I$ if and only if $f(S_I) \neq \bot$ and $f(S_I) \prec T$. Furthermore, $not \; f(S) \prec T$ is true (satisfied) with respect to $I$ if and only if $f(S_I) = \bot$ or $f(S_I) \neq \bot$ and $f(S_I) \nprec T$.
\end{definition}

\begin{definition} Let $\Pi$ be a ground NDLP program, $r$ be a ground NDLP rule of the form (\ref{rule:DLPN}), and $I$ be an interpretation. Then,
\begin{itemize}
\item $I$ satisfies $head(r)$ iff $\exists i$ $(1 \leq i \leq k)$ such that $I$ satisfies $a_i$.

\item $I$ satisfies $body(r)$ iff $\forall(k+1 \leq i \leq m)$ $I$ satisfies $a_i$ and $\forall(m+1 \leq j
\leq n)$ $I$ satisfies $not\; a_j$.

\item $I$ satisfies $r$ iff $I$ satisfies $head(r)$ whenever $I$ satisfies $body(r)$ or $I$ does not satisfy $body(r)$.

\item $I$ satisfies $\Pi$ iff $I$ satisfies every NDLP rule, $r$, in $\Pi$.
\end{itemize}
\end{definition}

\subsection{Answer Sets}

A model for an NDLP program, $\Pi$, is an interpretation that satisfies $\Pi$. A model $I$ of $\Pi$ is \emph{$\subseteq$-minimal} if and only if there does not exist a model $I'$ of $\Pi$ such that $I' \subset I$.

\begin{definition} Let $\Pi$ be a ground NDLP program, $r$ be an NDLP rule in $\Pi$, and $I$ be an interpretation. Let $I \models body(r)$ denotes that $I$ satisfies $body(r)$. Then, the reduct, $\Pi^I$, of $\Pi$ w.r.t. $I$ is the ground NDLP program $\Pi^I$ where
\[
\Pi^I = \{ head(r) \leftarrow body(r) \: \: | \: \: r \in \Pi \: \wedge \: I \models body(r)\}
\]
\end{definition}
The reduct $\Pi^I$ of $\Pi$ w.r.t. $I$ excludes all rules $r \in \Pi$ whose body, $body(r)$, is not satisfied by $I$. The satisfaction of $body(r)$ in the definition of the reduct does not distinguish between atoms or aggregate atoms or the negation of atoms or the negation aggregate atoms. This means that if dissatisfaction of $body(r)$ is due to unsatisfied atom or aggregate atom or unsatisfied negated atom or negated aggregate atom the consequence is the same, which is the exclusion of $r$ from the reduct of $\Pi$.

\begin{definition} An interpretation, $I$, for a ground NDLP program, $\Pi$, is an answer set for $\Pi$ if $I$ is $\subseteq$-minimal model for $\Pi^I$.
\end{definition}
Observe that the definitions of the reduct and the answer sets semantics for NDLP programs are generalizations of the definitions of the reduct and the answer sets semantics for a single aggregate over a variable disjunctive logic programs, DLP$^{\cal A}$, described in \cite{Recur-aggr}, and hence, generalizations of the definitions of the reduct and the answer sets semantics for the original disjunctive logic programs, DLP, presented in \cite{Gelfond_B}.

\subsection{Semantics Properties}

In this section we study the semantics properties of NDLP and its relationship to the answer set semantics of a single variable and a single aggregate disjunctive logic programs, DLP$^{\cal A}$ \cite{Recur-aggr}, and its relationship to the answer set semantics of the original disjunctive logic programs, DLP \cite{Gelfond_B}.

\begin{theorem} Let $\Pi$ be an NDLP program. The answer sets of $\Pi$ are $\subseteq$--minimal models for $\Pi$.
\end{theorem}
The following theorem shows that the answer set semantics of NDLP subsumes the answer set semantics of DLP$^{\cal A}$ \cite{Recur-aggr}, and consequently subsumes the original answer set semantics of the original disjunctive logic programs DLP \cite{Gelfond_B}. DLP$^{\cal A}$ programs are NDLP programs with a single aggregation over a single variable which, unlike NDLP programs, do not allow aggregations over function terms. DLP programs are NDLP programs without any aggregate atoms.

\begin{theorem} Let $\Pi$ be a DLP$^{\cal A}$ program and $I$ be an interpretation. Then, $I$ is an answer set for $\Pi$ iff $I$ is an answer set for $\Pi$ according to the answer set semantics of \cite{Recur-aggr}.
\end{theorem}

\begin{proposition} Let $\Pi$ be a DLP program and $I$ be an interpretation. Then, $I$ is an answer set for $\Pi$ iff $I$ is an answer set for $\Pi$ according to the answer set semantics of \cite{Gelfond_B}.
\end{proposition}

\section{Nested Aggregates Answer Set Optimization}

In this section, we introduce the syntax and semantics of the answer set optimization programs with preferences that involve nested multiple aggregates, called nested aggregates preferences, and denoted by NASO programs, that modify and generalize the syntax and semantics of the answer set optimization with aggregate preferences presented in \cite{Saad_ASOG}, from a single aggregate over a single variable preferences to nested multiple aggregates over multiple variables and nested multiple aggregates over functions that involve multiple variables preferences. An NASO program is a logic program under the answer set semantics whose answer sets are ranked according to preference relations  represented in the program.

An NASO program, $\Pi$, is a union of two sets of logic rules $\Pi =  R_{gen} \cup R_{pref}$. The first set of logic rules, $R_{gen}$, is called the generator rules that generate the answer sets that satisfy every rule in $R_{gen}$. $R_{gen}$ is any set of logic rules with well-defined answer set semantics including normal, extended, and disjunctive sets of rules \cite{Gelfond_A,Gelfond_B,Recur-aggr}, as well as disjunctive logic with nested multiple aggregates sets of rules presented in the first part of this paper. The second set of logic rules, $R_{pref}$, is a set of logic rules that represent the user preferences over the answer sets generated by $R_{gen}$, called the {\em preference rules}. The preference rules in $R_{pref}$ are used to rank the generated answer sets from the most preferred answer set to the least preferred one. An advantage of NASO is that $R_{gen}$ and $R_{pref}$ are independent. This makes preference elicitation easier and the whole approach is more intuitive and easy to use in practice. We focus on the syntax and semantics of the preference rules $R_{pref}$ of the NASO programs $\Pi = R_{gen} \cup R_{pref}$, since the syntax and semantics of $R_{gen}$ is the same as syntax and semantics of any set of logic rules
with answer set semantics \cite{Gelfond_B,Gelfond_A,Recur-aggr}.

\subsection{NASO Programs Syntax}

The language of NASO programs is the same as the language NDLP programs, presented in the first part of this paper, except that in the language of NASO programs classical negation is allowed. Let ${\cal L}$ be a first-order language with finitely many predicate symbols, function symbols, constants, and infinitely many variables. A term is a constant, a variable or a function. A literal is either an atom $a$ in $B_{\cal L}$ or the negation of $a$ ($\neg a$), where $B_{\cal L}$ is the Herbrand base of ${\cal L}$ and $\neg$ is the classical negation. The Herbrand universe of ${\cal L}$ is denoted by $U_{\cal L}$. Non-monotonic negation or the negation as failure is denoted by $not$. Let $Lit$ be the set of all literals in ${\cal L}$, where $Lit = \{a | a \in B_{\cal L}\} \cup \{\neg a | a \in B_{\cal L} \}$.

An expression of the form $\{ F \; | \; C \}$ is called a symbolic set, where $F$ is a single variable or a function $F = g(X_1, \ldots, X_n)$, where $X_1, \ldots, X_n$ are variables or functions terms and $C$ is a conjunction of literals, non-monotonic negation of literals, aggregate atoms, and non-monotonic negation of aggregate atoms (defined below). However, a ground set is a set of pairs of the form $\langle F_G \; | \; C_G \rangle$ such that $F_G$ is a constant and $C_G$ is a conjunction of ground literals, non-monotonic negation of ground literals, ground aggregate atoms, and non-monotonic negotiation of ground aggregate atoms (defined below). A ground set or a symbolic set is called a set term. We say $f(S)$ is an aggregate function if $f$ is an aggregate function symbol and $S$ is a set term, where $f \in \{ min, max, count, sum, times \}$. If $f(S)$ is an aggregate function and $T$ is a constant, variable or function term called guard, then we say $f(S) \prec T$ is an aggregate atom, where $\prec \in \{=, \neq, <, >, \leq, \geq \}$. An optimization aggregate is an expression of the form $max(f(S))$ or $min(f(S))$, where $S$ is a set term and $f$ is an aggregate function symbol. Let ${\cal A}$ be a set of literals, aggregate atoms, and optimization aggregates. A boolean combination over ${\cal A}$ is a boolean formula over literals, aggregates atoms, and optimization aggregates in ${\cal A}$ constructed by conjunction, disjunction, and non-monotonic negation ($not$), where non-monotonic negation is combined only with literals and aggregate atoms.

Similar to NDLP, an aggregate function $f(S)$ in an aggregate atom $f(S) \prec T$ can be defined in terms of other  aggregate atoms $f'(S') \prec T'$ and hence in terms of other aggregate functions $f'(S')$. This is because the conjunction in the definition of the symbolic set $S$ of the aggregate function $f(S)$ in the aggregate atom $f(S) \prec T$ can contain other aggregate atoms as constituents. This implies that multiple level of nested aggregate functions calculation is achieved by allowing aggregate functions of the aggregate atoms to be defined in terms of other aggregate atoms.

Let $f(S)$ be an aggregate function. A variable, $X$, is a local variable to $f(S)$ if and only if $X$ appears in $S$ and $X$ does not appear in any aggregate function that is outer to $f(S)$ or in the preference rule the contains $f(S)$. A global variable is a variable that is not a local variable.

\begin{definition} A preference rule, $r$, over a set of literals, aggregate atoms, and optimization aggregates, ${\cal A}$, is an expression of the form
\begin{equation}
C_1 \succ C_2 \succ \ldots \succ C_k \leftarrow l_{k+1},\ldots, l_m, not\; l_{m+1},\ldots, not\;l_{n} \label{rule:pref}
\end{equation}
where $l_{k+1}, \ldots, l_{n}$ are literals or aggregate atoms and $C_1, C_2, \ldots, C_k$ are boolean combinations over ${\cal A}$.
\end{definition}
Let $r$ be a preference rule, $body(r) = l_{k+1},\ldots, l_m, not\; l_{m+1},\ldots, not\;l_{n}$, and $head(r) = C_1 \succ C_2 \succ \ldots \succ C_k$. Intuitively, a preference rule, $r$, says that any answer set that satisfies $body(r)$ and $C_1$ is preferred over answer sets that satisfy $body(r)$, some $C_i$ $(2 \leq i \leq k)$, but not $C_1$, and any answer set that satisfies $body(r)$ and $C_2$ is preferred over answer sets that satisfy $body$, some $C_i$ $(3 \leq i \leq k)$, but neither $C_1$ nor $C_2$, etc.

Recalling, an NASO program is a union of two sets of logic rules $\Pi =  R_{gen} \cup R_{pref}$, where $R_{gen}$ is a set of logic rules with answer set semantics, called the generator rules, and $R_{pref}$ is a set of preference rules.

\subsection{NASO Programs Semantics}

In defining the semantics of aggregate functions and aggregate atoms for NASO programs as they syntactically defined in the previous section, we follow the same semantics of aggregate functions and aggregate atoms as they defined for NDLP programs. We use for NASO programs the same notions of local ground instantiation and ground instantiation of symbolic sets as they defined for NDLP programs. Similarly, the semantics of the aggregate functions $min$, $max$, $count$, $sum$, and $times$ are defined by the same mappings as defined for NDLP programs. Let $\bot$ be a symbol that does not occur in any NASO program.

A ground instantiation of a preference rule, $r$, is the replacement of each global variable appearing in $r$ to a constant from $U_{\cal L}$, then followed by the ground instantiation of every symbolic set, $S$, appearing in $r$. The ground instantiation of an NASO program, $\Pi = R_{gen} \cup R_{pref}$, is the set of all possible ground instantiations of every rule, $r$, in $\Pi$. Let $I$ be an answer set for $R_{gen}$ in an NASO program, $\Pi = R_{gen} \cup R_{pref}$, $f(S) \prec T$ be a ground aggregate atom, and $S_I = \{\!\!\{F_G \; | \; \langle F_G \: | \: C_G \rangle \in S$ and $C_G$ is true in $I \}\!\!\}$ be the multiset constructed from elements in $S$. Then, the evaluation of $f(S)$ with respect to the answer set $I$ is $f(S_I)$, where $f(S_I) = \bot$ if $S_I$ is not in the domain of $f$.

\begin{definition} Let $\Pi = R_{gen} \cup R_{pref}$ be a ground NASO program, $I$ be an answer set for $R_{gen}$, and $r$ be a preference rule in $R_{pref}$. Then the satisfaction of a boolean combination, $C$, appearing in $head(r)$, by $I$, denoted by $I \models C$, is defined inductively as follows:

\begin{itemize}

\item $I \models l$ iff  $l \in I$.

\item $I \models not \; l$ iff  $l \notin I$.

\item $I \models f(S) \prec T$ iff  $f(S_I) \neq \bot$ and $f(S_I) \prec T$.

\item $I \models not \; f(S) \prec T$ iff $f(S_I) = \bot$ or $f(S_I) \neq \bot$ and $f(S_I) \nprec T$.

\item $I \models max(f(S))$ iff $f(S_I) \neq \bot$, and for any answer set $I'$, $f(S_{I'}) \neq \bot$ and $f(S_{I'}) \leq f(S_{I})$
or $f(S_I) \neq \bot$ and $f(S_{I'}) = \bot$.

\item $I \models min(f(S))$ iff $f(S_I) \neq \bot$, and for any answer set $I'$, $f(S_{I'}) \neq \bot$ and $f(S_{I}) \leq f(S_{I'})$
or $f(S_I) \neq \bot$ and $f(S_{I'}) = \bot$.

\item $I \models C_1 \wedge C_2$ iff $I \models C_1 $ and $I \models C_2$.

\item $I \models C_1 \vee C_2$ iff $I \models C_1 $ or $I \models C_2$.

\end{itemize}
\end{definition}
The application of any aggregate function, $f$, except $count$, on a singleton $\{x\}$, returns $x$, i.e., $f(\{x\}) = x$. Therefore, we use $max(S)$ and $min(S)$ as abbreviations for the optimization aggregates $max(f(S))$ and $min(f(S))$ respectively, where $S$ is a singleton and $f$ is any arbitrary aggregate function except $count$.

\begin{definition} Let $\Pi = R_{gen} \cup R_{pref}$ be a ground NASO program, $I$ be an answer set for $R_{gen}$, and $r$ be a preference rule in $R_{pref}$. Then the satisfaction of the body of $r$ by $I$, denoted by $I \models body(r)$, is defined inductively as follows:

\begin{itemize}

\item $I \models l$ iff  $l \in I$.

\item $I \models not \; l$ iff  $l \notin I$.

\item $I \models f(S) \prec T$ iff  $f(S_I) \neq \bot$ and $f(S_I) \prec T$.

\item $I \models not \; f(S) \prec T$ iff $f(S_I) = \bot$ or $f(S_I) \neq \bot$ and $f(S_I) \nprec T$.

\item $I \models body(r)$ iff $\forall (k+1 \leq i \leq m)$, $I \models l_i$, and $\forall (m+1 \leq j \leq n)$, $I \models not \; l_j$.

\end{itemize}
\end{definition}
The following definition specifies the satisfaction of the preference rules.

\begin{definition} Let $\Pi = R_{gen} \cup R_{pref}$ be a ground NASO program, $I$ be an answer set for $R_{gen}$, $r$ be a preference rule in $R_{pref}$, and $C_i$ be in $head(r)$. Then, we define the following notions of satisfaction of $r$ by $I$:
\begin{itemize}
\item $I \models_{i} r$ iff $I \models body(r)$ and $I \models C_i$.

\item $I \models_{irr} r$ iff $I \models body(r)$ and $I$ does not satisfy any of $C_i$ in $head(r)$.

\item $I \models_{irr} r$ iff $I$ does not satisfy $body(r)$.
\end{itemize}
\end{definition}

\begin{definition} Let $\Pi = R_{gen} \cup R_{pref}$ be a ground NASO program, $I_1, I_2$ be two answer sets of $R_{gen}$, $r$ be a preference rule in $R_{pref}$, and $C_l$ be boolean combination appearing in $head(r)$. Then, $I_1$ is strictly preferred over $I_2$ w.r.t. $r$, denoted by $I_1 \succ_r I_2$, iff one of the following holds:
\begin{itemize}
\item $I_1 \models_{i} r$ and $I_2 \models_{j} r$ and $i < j$, \\
where $i = \min \{l \; | \; I_1 \models_l r \}$ and $j = \min \{l \; | \; I_2 \models_l r \}$.

\item $I_1 \models_{i} r$ and $I_2 \models_{irr} r$.
\end{itemize}
We say, $I_1$ and $I_2$ are equally preferred w.r.t. $r$, denoted by $I_1 =_{r} I_2$, iff one of the following holds:
\begin{itemize}
\item $I_1 \models_{i}  r$ and $I_2 \models_{i} r$,
where $i = \min \{l \; | \; I_1 \models_l r \} = \min \{l \; | \; I_2 \models_l r \}$.
\item $I_1 \models_{irr}  r$ and $I_2 \models_{irr} r$.
\end{itemize}
We say, $I_1$ is at least as preferred as $I_2$ w.r.t. $r$, denoted by $I_1 \succeq_{r} I_2$, iff $I_1 \succ_{r} I_2$ or $I_1 =_{r} I_2$.
\label{def:pref-rule}
\end{definition}
Definition (\ref{def:pref-rule}) specifies the ranking of the answer sets according to a preference rule. The following definitions characterize the ranking of the answer sets with respect to a set of preference rules.

\begin{definition} [Pareto Preference] Let $\Pi = R_{gen} \cup R_{pref}$ be an NASO program and $I_1, I_2$ be answer sets of $R_{gen}$. Then, $I_1$ is (Pareto) preferred over $I_2$ w.r.t. $R_{pref}$, denoted by $I_1 \succ_{R_{pref}} I_2$, iff there exists at least one preference rule $r \in R_{pref}$ such that $I_1 \succ_{r} I_2$ and for every other rule $r' \in R_{pref}$, $I_1 \succeq_{r'} I_2$. We say, $I_1$ and $I_2$ are equally (Pareto) preferred w.r.t. $R_{pref}$, denoted by $I_1 =_{R_{pref}} I_2$, iff for all $r \in R_{pref}$, $I_1 =_{r} I_2$.
\end{definition}

\begin{definition} [Maximal Preference] Let $\Pi = R_{gen} \cup R_{pref}$ be an NASO program and $I_1, I_2$ be answer sets of $R_{gen}$. Then, $I_1$ is (Maximal) preferred over $I_2$ w.r.t. $R_{pref}$, denoted by $I_1 \succ_{R_{pref}} I_2$, iff
\[
|\{r \in R_{pref} | I_1 \succeq_{r} I_2\}| > |\{r \in R_{pref} | I_2 \succeq_{r} I_1\}|.
\]
We say, $I_1$ and $I_2$ are equally (Maximal) preferred w.r.t. $R_{pref}$, denoted by $I_1 =_{R_{pref}} I_2$, iff
\[
|\{r \in R_{pref} | I_1 \succeq_{r} I_2\}| = | \{r \in R_{pref} | I_2 \succeq_{r} I_1\}|.
\]
\end{definition}
Observe that the Maximal preference relation is more {\em general} than the Pareto preference relation, since the Maximal preference definition {\em subsumes} the Pareto preference relation. Under the Pareto preference relation, the following result shows that the syntax and semantics of NASO programs subsume the syntax and semantics of the answer set optimization programs of \cite{ASO}, since there is no notion of Maximal preference relation was introduced in \cite{ASO}. This is assuming that the answer set optimization programs of \cite{ASO} assign the lowest rank to the answer sets that do not satisfy neither the body nor the head of preference rules.

\begin{theorem}
Let $\Pi = R_{gen} \cup R_{pref}$ be an NASO program without either aggregate atoms or optimization aggregates and $I_1, I_2$ be answer sets of $R_{gen}$. Then, $I_1$ is Pareto preferred over $I_2$ w.r.t. $R_{pref}$ iff $I_1$ is Pareto preferred over $I_2$ w.r.t. $R_{pref}$ according to \cite{ASO}.
\end{theorem}
However, the following result shows that the syntax and semantics of NASO programs subsume the syntax and semantics of a single aggregate over a single variable answer set optimization programs of \cite{Saad_ASOG}, under both the Pareto and the Maximal preference relations, since the notion of Maximal preference relation introduced in \cite{Saad_ASOG} is a special case of the Maximal preference relation presented in this paper.

\begin{theorem} Let $\Pi = R_{gen} \cup R_{pref}$ be a single aggregate over a single variable NASO program and $I_1, I_2$ be answer sets of $R_{gen}$. Then, $I_1$ is Pareto (Maximal) preferred over $I_2$ w.r.t. $R_{pref}$ iff $I_1$ is Pareto (Maximal) preferred over $I_2$ w.r.t. $R_{pref}$ according to \cite{Saad_ASOG}.
\end{theorem}

\section{Probabilistic Traveling Salesman Problem}

In this section we show that any instance of {\em a priori} optimization probabilistic traveling salesman problem (PTSP) \cite{PTSP} can be intuitively and easily represented and solved by the framework of nested multiple aggregates answer set optimization programs. In addition, we show that with a minor modification to our representation of any instance of PTSP problem in NASO, we can intuitively solve a corresponding classical traveling salesman problem (TSP) in our framework in particular, and in answer set programming in general, since finding the optimal tour for TSPs has not been considered before in answer set programming literature. This is because the emphasis in solving TSPs in answer set programming literature was on generating the possible tours of a given TSP rather than finding the optimal tour for that TSP, which is the tour with the minimum length. The reason for that is tours of a TSP are represented in answer set programming as answer sets and answer set programming is incapable of reasoning across answer sets to find the answer set that represents the tour with the minimum length. Therefore, a different framework that is capable of reasoning across answer sets is required, this framework is NASO programs.

Observe that our representation of PTSPs and consequently of TSPs is built on top of the existing answer set programming representation of TSPs, whose aim is to find all possible tours. This makes our NASO program representation further intuitive and keeps inline with the existing body of work in answer set programming. In addition, it shows and gives insight that many optimization problems can be solved by NASO framework in the same way by similar intuitive modifications to the existing answer set programming representation of corresponding similar problems.

\subsection{Probabilistic Traveling Salesman Problem in NASO}

Recalling, a Probabilistic Traveling Salesman Problem (PTSP), $G$, is a tuple of the form $G = \langle V, E, D, P \rangle$, where $V$ is a set of $n$ vertices, $E$ is a set of edges, $D$ is an $n \times n$ distance matrix, where $d_{ij}$ is distance from a vertex $i$ to a vertex $j$, and $P = \{p_1, p_2, \ldots, p_n\}$ is a probability distribution over vertices in $V$, where $p_i$ is the probability a vertex $i$ must be visited. The aim is to find a {\em priori a tour}, $t = (0, 1, 2, 3, 4,  \ldots, n, 0 )$, of $G$, with the minimum expected length, where $0$ is the starting point from which the tour always starts and ends and whose probability $p_0 = 1$. The expected length, $E[L_t]$, of a tour $t$ is given by formula (\ref{eq:expected}).

Any Probabilistic Traveling Salesman Problem (PTSP), $G =  \langle V, E, D, P \rangle$, is represented as an NASO program $\Pi = R_{gen} \cup R_{pref}$, where the generator rules in $R_{gen}$ generate the answer sets the represents all the possible tours in $G$ and the preference rules in $R_{pref}$ rank the answer sets that correspond to all the possible tours in $G$ from the tour with the minimum expected length to the tour with the maximum expected length. The representation proceeds as follows.
\begin{itemize}
\item Every vertex $x$ in $V$ whose probability is $p$ in $P$ is represented in $R_{gen}$ as a fact of the form
\begin{equation}
vertex(x,p) \leftarrow \label{rule:vtx}
\end{equation}
In addition to the starting vertex $y \in V$ is represented in $R_{gen}$ as a fact of the form
\begin{equation}
start(y) \leftarrow
\end{equation}
\item Every edge $(x, y)$ in $E$ is represented in $R_{gen}$ as a fact of the form
\begin{equation}
edge(x,y) \leftarrow
\end{equation}

\item Every element $d_{xy}$ in the distance matrix $D$ is represented in $R_{gen}$ as a fact of the form
\begin{equation}
distance(x,y, d_{xy}) \leftarrow
\end{equation}

\item The logic rules that generate all the possible tours from, $G$, are represented in $R_{gen}$ by the disjunctive logic rules:
\begin{eqnarray}{r}
inTour(X,Y) \vee outTour(X,Y)  \leftarrow & start(X), \notag \\ edge(X,Y). \label{rule:in1} \\
inTour(X,Y) \vee outTour(X,Y)  \leftarrow & reached(X), \notag \\ edge(X,Y) \label{rule:in2}. \\
reached(Y)  \leftarrow  inTour(X,Y) \label{rule:reach}.
\end{eqnarray}
\begin{eqnarray}
 \leftarrow  inTour(X,Y), inTour(X,Y_1), Y \neq Y_1. \label{rule:constr1}\\
 \leftarrow  inTour(X,Y), inTour(X_1,Y), X \neq X_1. \label{rule:constr2} \\
 \leftarrow  vertex(X, P), not \; reached(X). \label{rule:constr3} \\
\end{eqnarray}

\item The indexing of vertices in any given tour, $t$, by their order of appearance in the tour $t$ is represented in $R_{gen}$ by the logic rules:
\begin{eqnarray}
index(0, X)  \leftarrow   start(X). \label{rule:index1} \\
index(I+1, Y)  \leftarrow  inTour(X,Y), edge(X,Y), \notag \\ index(I, X) \label{rule:index2}.
\end{eqnarray}

\item The computation of the expected length of each tour and the ranking of the answer sets corresponding to the tours in $G$ from the minimum expected length tour to the maximum expected length tour is represented in $R_{pref}$ by the preference rule (denoted by the preference rule $opt$):
\end{itemize}
\[
\begin{array}{c}
min (\qquad sum \; \{ \qquad A \qquad| \;
\\
index(I, X), \; vertex(X, P_X), \;  0 \leq I \leq  n, \;
\\ sum \; \{ \qquad P_X * P_Y * D * M \qquad | \qquad
\\
index(J, Y), \; vertex(Y, P_Y), \; distance(X, Y, D), \\ I + 1 \leq  J \leq  n + 1, \;
\\ times \; \{ \qquad (1 - P_Z) \qquad | \;
\\
index(K, Z), \; vertex(Z, P_Z),\; I + 1 \leq K \leq  J - 1 \} = M \; \} \;  = A \; \} \\ \qquad ) \leftarrow
\end{array}
\]
Observe that the logic rules (\ref{rule:in1})--(\ref{rule:reach}) are exactly a typical set of logic rules that are used in answer set programming literature \cite{DLV} to find Hamiltonian cycles of a given graph adapted to deal with graphs with probabilistic vertices. It can be easily seen that the nested aggregate function within the optimization aggregate $min$ in the above preference rule is exactly the representation of the expected length of a tour described by formula (\ref{eq:expected}).

\begin{theorem} Let $G = \langle V, E, D, P \rangle$ be a Probabilistic Traveling Salesman Problem and $\Pi = R_{gen} \cup R_{pref}$ be the NASO program representation of $G$. Then, a tour, $t$, is the optimal tour for $G$ with the minimum expected length iff an answer set, $I$, of $R_{gen}$ is the top preferred answer set w.r.t. $R_{pref}$.
\end{theorem}

\subsection{Traveling Salesman Problem in NASO}

In this section, we show that with a minor modification to the NASO program representation of a probabilistic traveling salesman problem, we can intuitively solve a corresponding classical traveling salesman problem (TSP). A classical Traveling Salesman Problem (TSP), $G$, is a tuple of the form $G = \langle V, E, D \rangle$, where $V$ is a set of $n$ vertices, $E$ is a set of edges, and $D$ is an $n \times n$ distance matrix, where $d(i,j)$ is an element in $D$ that represents the distance from a vertex $i$ to a vertex $j$. The aim is to find a tour, $t = (0, 1, 2, 3, 4,  \ldots, n, 0 )$, of $G$, with the minimum length, where $0$ is the starting point from which the tour always starts and ends and the length, $L_t$, of a tour $t$ is calculated by
\begin{equation}
L_t = \sum_{i = 0}^{n}  d(i,i+1)
\label{eq:length}
\end{equation}
where $n + 1 = 0$. A Traveling Salesman Problem (TSP), $G =  \langle V, E, D \rangle$, is represented as an NASO program, $\Pi = R_{gen} \cup R_{pref}$, where the generator rules in $R_{gen}$ generate the answer sets the represents all the possible tours in $G$ and the preference rules in $R_{pref}$ rank the answer sets that correspond to all the possible tours in $G$ from the tour with the minimum length to the tour with the maximum length. The
generator rules in $R_{gen}$ consists of the logic rules from the logic rule (\ref{rule:vtx}) through the logic rule (\ref{rule:index2}) after replacing $vertex(x,p)$ in the logic rule (\ref{rule:vtx}) by $vertex(x)$ and replacing $vertex(X,P)$ in the logic rule (\ref{rule:constr3}) by $vertex(X)$. The computation of the length of each tour and the ranking of the answer sets corresponding to the tours in $G$ from the minimum length tour to the maximum length tour is represented in $R_{pref}$ by the preference rule:
\[
\begin{array}{c}
min ( \; sum \; \{ \; D \; | \;
distance(X, Y, D), \; index(I, X),\; \\ index(I+1, Y), \;  0 \leq I \leq  n \;   \;  \} \; ) \leftarrow
\end{array}
\]

\begin{theorem} Let $G = \langle V, E, D \rangle$ be a classical Traveling Salesman Problem and $\Pi = R_{gen} \cup R_{pref}$ be the NASO program representation of $G$. Then, a tour, $t$, is the optimal tour for $G$ with the minimum length iff an answer set, $I$, of $R_{gen}$ is the top preferred answer set w.r.t. $R_{pref}$.
\end{theorem}

\subsection{Probabilistic Traveling Salesman Problem in NDLP}

We also show that a Probabilistic Traveling Salesman Problem, $G = \langle V, E, D, P \rangle$, can be represented as an NDLP program, $\Pi$, whose answer sets corresponds to tours in $G$. Although, NDLP programs framework is still capable of finding all possible tours in $G$, represented as answer sets, and computing the expected length of each tour, the semantics of NDLP programs is not able to locate the answer set that corresponds to the optimal tour of $G$, since it does not have the capability of reasoning across the answer sets. However, PTSPs can be represented and solved in NDLP programs framework in two steps. The first step is to represent a probabilistic traveling salesman problem, $G$, by an NDLP program whose answer sets correspond to all possible tours in $G$ along with their expected length. The second step is to determine the minimum expected length optimal tour, $t$, in $G$ represented by an answer set of the NDLP program representation of $G$ by means of any appropriate procedure internal or external to NDLP programs framework.

The NDLP program, $\Pi$, representation of a Probabilistic Travelling Salesman Problem, $G$, consists of the NDLP rules from the logic rule (\ref{rule:vtx}) through the logic rule (\ref{rule:index2}), in addition to the following NDLP rule, where $X$ in $length(X)$ in the NDLP rule below represents the value of the expected length of a tour.
\[
\begin{array}{c}
length(X) \quad \leftarrow  \quad sum \; \{ \; A \; | \;
\\
index(I, X), \; vertex(X, P_X), \;  0 \leq I \leq  n, \;
\\ sum \; \{ \; P_X * P_Y * D * M \; | \;
\\
index(J, Y), \; vertex(Y, P_Y), \; distance(X, Y, D), \; \\ I + 1 \leq  J \leq  n + 1, \;
\; times \; \{ \; (1 - P_Z) \; | \;
\\
index(K, Z), \; vertex(Z, P_Z),\; I + 1 \leq K \leq  J - 1 \} = M \; \} \\ \; = A \; \}  \quad = \quad X.
\end{array}
\]
\begin{theorem} Let $G = \langle V, E, D, P \rangle$ be a Probabilistic Traveling Salesman Problem and $\Pi$ be the NDLP program representation of $G$. Then, $t$ is a tour for $G$ iff $I$ is an answer set for $\Pi$ corresponds to $t$, where the expected length, $E[L_t]$, of $t$ is equal to the value of $X$ in $length(X)$ that is satisfied by $I$.
\end{theorem}

One possible way to find the optimal tour with the minimum expected length is to add the following two NDLP rules to the NDLP program representation, $\Pi$, of a probabilistic traveling salesman problem problem, $G$, where $X$ in $min(X)$ in the NDLP rules below represents the minimum expected length value of a tour
\begin{eqnarray}
min(X) \leftarrow  length(X), not \; maximal(X).  \label{rule:min1} \\
maximal(X) \leftarrow  length(X), length(X_1), X_1 < X. \label{rule:min2}
\end{eqnarray}

\begin{theorem} Let $G = \langle V, E, D, P \rangle$ be a Probabilistic Traveling Salesman Problem and $\Pi$ be the NDLP program representation of $G$. Then, a tour, $t$, is the optimal tour for $G$ with the minimum expected length iff an answer set $I$ of $\Pi$ is the only answer set of $\Pi$ that satisfies $min(X)$.
\end{theorem}
However, the addition of the NDLP rules (\ref{rule:min1}) and (\ref{rule:min2}) to the NDLP program representation of a probabilistic traveling salesman problem leads when grounded to large number of ground rules, especially with the graphs that involve sufficiently large number of vertices and large number of edges, which is always the case with PTSPs. Therefore, a simple linear search for the minimum value of $X$ in $length(X)$ over all the answer sets of the NDLP program representation of a probabilistic traveling salesman problem is likely to be more efficient than performing the same task using logic rules.

\subsection{Working Example}

This section shows that a probabilistic traveling salesman problem instance can be intuitively and easily represented and solved by the NASO programs framework. Consider this instance of the probabilistic traveling salesman problem, $G = \langle V, E, D, P \rangle$, where $\langle V, E \rangle$ is complete undirected graph where $V = \{a, b, c, d\}$ and $E = \{
(a,b),
(a,c),
(a,d),
(b,c),
(b,d),
(c,d),
(b,a),
(c,a),
(d,a), \\
(c,b),
(d,b),
(d,c)\}$, the probability distribution, $P$, is given as $p_a = 1$, $p_b = 0.3$, $p_c = 0.7$, and $p_d = 0.4$, and the distance matrix, $D$, is given by
$d_{ab} = 40$,
$d_{ac} = 40$,
$d_{ad} = 22$,
$d_{bc} = 40$,
$d_{bd} = 25$,
$d_{cd} = 22$,
$d_{ba} = 40$,
$d_{ca} = 40$,
$d_{da} = 22$,
$d_{cb} = 40$,
$d_{db} = 25$, and
$d_{dc} = 22$.

Let $\Pi = R_{gen} \cup R_{pref}$ be the NASO program representation of this probabilistic traveling salesman problem instance, $G$, where $R_{pref}$ consists of the preference rule $opt$, with $n = 3$, and $R_{gen}$ consists of the logic rules (\ref{rule:in1}) through (\ref{rule:index2}), in addition to the fats:
\[
\begin{array}{c}
vertex(a, 1). \quad
vertex(b, 0.3). \quad
vertex(c, 0.7). \quad
vertex(d, 0.4).
\\
edge(a,b). \quad
edge(a,c). \quad
edge(a,d). \quad
edge(b,c). \quad
edge(b,d).
\\
edge(c,d). \quad
edge(b,a). \quad
edge(c,a). \quad
edge(d,a). \quad
edge(c,b).
\\
edge(d,b). \quad
edge(d,c). \quad
distance(a,c, 40). \quad
distance(a,b, 40).
\\
distance(a,d, 22). \quad
distance(b,c, 40). \quad
distance(b,d, 25).
\\
distance(c,d, 22). \quad
distance(c,a, 40). \quad
distance(b,a, 40).
\\
distance(c,b, 40). \quad
distance(d,b, 25). \quad
distance(d,a, 22).
\\
distance(d,c, 22). \quad
start(a).
\end{array}
\]
By considering only the relevant atoms, $R_{gen}$, has six answer sets which correspond to the six available tours in $G$. These answer sets are:
\[
\begin{array}{l}
I_1 = \{
start(a), \;
inTour(a,d), \;
inTour(d,b), \;
inTour(b,c),
\\
inTour(c,a), \;
index(0,a), \;
index(1,d), \;
index(2,b), \\
index(3,c), \;
index(4,a)
\}
\end{array}
\]
\[
\begin{array}{l}
I_2 = \{
start(a), \;
inTour(a,d), \;
inTour(d,c), \;
inTour(c,b),
\\
inTour(b,a), \;
index(0,a), \;
index(1,d), \;
index(2,c),
\\
index(3,b), \;
index(4,a)
\}
\end{array}
\]
\[
\begin{array}{l}
I_3 =  \{
start(a), \;
inTour(a,b), \;
inTour(b,d), \;
inTour(d,c),
\\
inTour(c,a), \;
index(0,a), \;
index(1,b), \;
index(2,d),
\\
index(3,c), \;
index(4,a)
\}
\end{array}
\]
\[
\begin{array}{l}
I_4 = \{
start(a), \;
inTour(a,c), \;
inTour(c,b), \;
inTour(b,d),
\\
inTour(d,a), \;
index(0,a), \;
index(1,c), \;
index(2,b),
\\
index(3,d), \;
index(4,a)
\}
\end{array}
\]
\[
\begin{array}{l}
I_5 =  \{
start(a), \;
inTour(a,c), \;
inTour(c,d), \;
inTour(d,b),
\\
inTour(b,a), \;
index(0,a), \;
index(1,c), \;
index(2,d),
\\
index(3,b), \;
index(4,a)
\}
\end{array}
\]
\[
\begin{array}{l}
I_6 = \{
start(a), \;
inTour(a,b), \;
inTour(b,c), \;
inTour(c,d),
\\
inTour(d,a), \;
index(0,a), \;
index(1,b), \;
index(2,c), \;
\\
index(3,d), \;
index(4,a)
\}
\end{array}
\]
There are six tours for, $G$, which are $t_1$, $t_2$, $t_3$, $t_4$, $t_5$, and $t_6$ that correspond to the six answer sets $I_1$, $I_2$, $I_3$, $I_4$, $I_5$, and $I_6$ respectively. The expected length, $E[L_{t_i}]$, of each tour, $t_i$, in $G$, calculated by formula (\ref{eq:expected}), is given as:
\[
\begin{array}{c}
E[L_{t_1}] =  75.16, \quad
E[L_{t_2}] = 76.67, \quad
E[L_{t_3}] = 76.92, \\
E[L_{t_4}] = 76.92, \quad
E[L_{t_5}] = 76.92, \quad
E[L_{t_6}] = 76.67
\end{array}
\]
that exactly corresponds to the evaluation of the nested aggregate function within the optimization aggregate, $min$, of the preference rule $opt$ contained in the set of the preference rules $R_{pref}$ of the NASO program representation, $\Pi = R_{gen} \cup R_{pref}$, of the probabilistic traveling salesman problem instance $G$. It is clear that $t_1$ is the tour with minimum expected length, $E[L_{t_1}] =  75.16$, and hence the optimal tour for $G$, as well as, $I_1$ is the top answer set of $R_{gen}$ with respect to $R_{pref}$. This is because
\[
\begin{array}{c}
I_1 \models_{1} opt,  \quad I_2 \models_{irr} opt,  \quad  I_3 \models_{irr} opt,  \quad  I_4 \models_{irr} opt,
\\ I_5 \models_{irr} opt,  \quad I_6 \models_{irr} opt.
\end{array}
\]
Similarly, this probabilistic traveling salesman problem instance can be represented by an NDLP program, $\Pi'$, that consists of the NDLP rules (\ref{rule:in1}) through (\ref{rule:min2}), as well as the NDLP rule representation of the expected length (the definition of the $length(X)$ predicate), and by considering only the relevant atoms, it can be easily seen that $I_1$ is the only answer set of $\Pi'$ that satisfies $min(X)$ and coincides with the optimal minimum expected length tour of $G$.

\section{Conclusions and Related Work}

We developed a generalization for the syntax and semantics of both answer set optimization programming and answer set programming from a single aggregate over a single variable to nested multiple aggregates over multiple variables and to nested multiple aggregates over functions involving multiple variables. In addition, we showed the applicability of both the nested aggregates answer set optimization programming and the nested aggregates answer set programming to the {\em a priori} optimization Probabilistic Traveling Salesman Problem, a fundamental stochastic optimization problem in Operation Research. Moreover, we showed that with minor modification to the probabilistic traveling salesman problems representation in nested aggregate answer set optimization programming and answer set optimization programming, a corresponding classical traveling salesman problems can be intuitively represented and solved in both frameworks.

We showed that the nested aggregates answer set optimization programming framework presented in this paper modifies and generalizes a single aggregate over a single variable answer set optimization programming framework of \cite{Saad_ASOG} as well as a generalization of the answer set optimization programming framework of \cite{ASO}. We showed that the nested aggregates answer set programming framework presented in this paper subsumes a single aggregate over a single variable answer set programming framework of \cite{Recur-aggr} as well as a generalization of the original answer set programming framework of \cite{Gelfond_B}.

The major difference in the development presented in this paper is that we allow the ability to reasoning in the presence of nested multiple aggregates in answer sets in general and in both answer set optimization programming and in answer set programming in particular. The only answer set optimization framework that allows aggregates is \cite{Saad_ASOG}, which is answer set optimization programs with a single aggregate over a single variable. The existing answer set programming literature considered a single aggregate over a single variable answer set programming approaches. A comprehensive comparisons of the existing a single aggregate over a single variable answer set programming approaches is found in \cite{Recur-aggr}.

\bibliographystyle{named}
\bibliography{Saad13NAASP}

\end{document}